\documentclass[10pt,twocolumn,letterpaper]{article}

\usepackage{iccv}
\usepackage{times}
\usepackage{epsfig}
\usepackage{graphicx}
\usepackage{amsmath}
\usepackage{amssymb}
\usepackage{bbm}

\usepackage[numbers]{natbib}
\usepackage[font=small,skip=0pt]{caption}
\usepackage{subcaption}
\usepackage{xcolor}
\usepackage{bm}
\usepackage{amsfonts}
\usepackage{enumitem}
\usepackage{soul}
\usepackage{array,makecell}
\usepackage[pagebackref=true,breaklinks=true,colorlinks,bookmarks=false]{hyperref}

\iccvfinalcopy

\begin{document}

\title{Always Be Dreaming: A New Approach for\\Data-Free Class-Incremental Learning}

\author{James Smith\textsuperscript{1}\thanks{Correspondence to: James Smith jamessealesmith@gatech.edu}, Yen-Chang Hsu\textsuperscript{2}, Jonathan Balloch\textsuperscript{1}, Yilin Shen\textsuperscript{2}, Hongxia Jin\textsuperscript{2}, Zsolt Kira\textsuperscript{1} \\
\normalsize
\textsuperscript{1}Georgia Institute of Technology,
\textsuperscript{2}Samsung Research America
}

\maketitle

\begin{abstract}
    Modern computer vision applications suffer from catastrophic forgetting when incrementally learning new concepts over time. The most successful approaches to alleviate this forgetting require extensive replay of previously seen data, which is problematic when memory constraints or data legality concerns exist. In this work, we consider the high-impact problem of \textit{Data-Free Class-Incremental Learning} (DFCIL), where an incremental learning agent must learn new concepts over time without storing generators or training data from past tasks. One approach for DFCIL is to replay synthetic images produced by inverting a frozen copy of the learner's classification model, but we show this approach fails for common class-incremental benchmarks when using standard distillation strategies. We diagnose the cause of this failure and propose a novel incremental distillation strategy for DFCIL, contributing a modified cross-entropy training and importance-weighted feature distillation, and show that our method results in up to a 25.1\% increase in final task accuracy (absolute difference) compared to SOTA DFCIL methods for common class-incremental benchmarks. Our method even outperforms several standard replay based methods which store a coreset of images. Our code is available at \url{https://github.com/GT-RIPL/AlwaysBeDreaming-DFCIL}
\end{abstract}

\begin{figure*}[t]
    \centering
    \begin{subfigure}[t]{0.32\textwidth}
        \centering
       
        \includegraphics[width=1\textwidth]{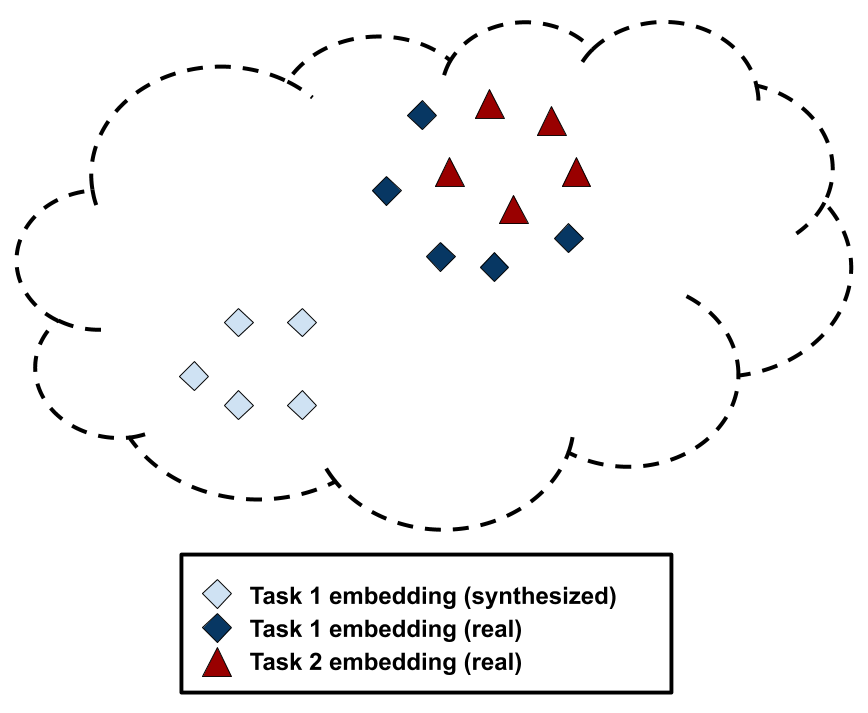}
         \caption{}
    \end{subfigure}
    \hfill
    \begin{subfigure}[t]{0.32\textwidth}
        \centering
        
        \includegraphics[width=1\textwidth]{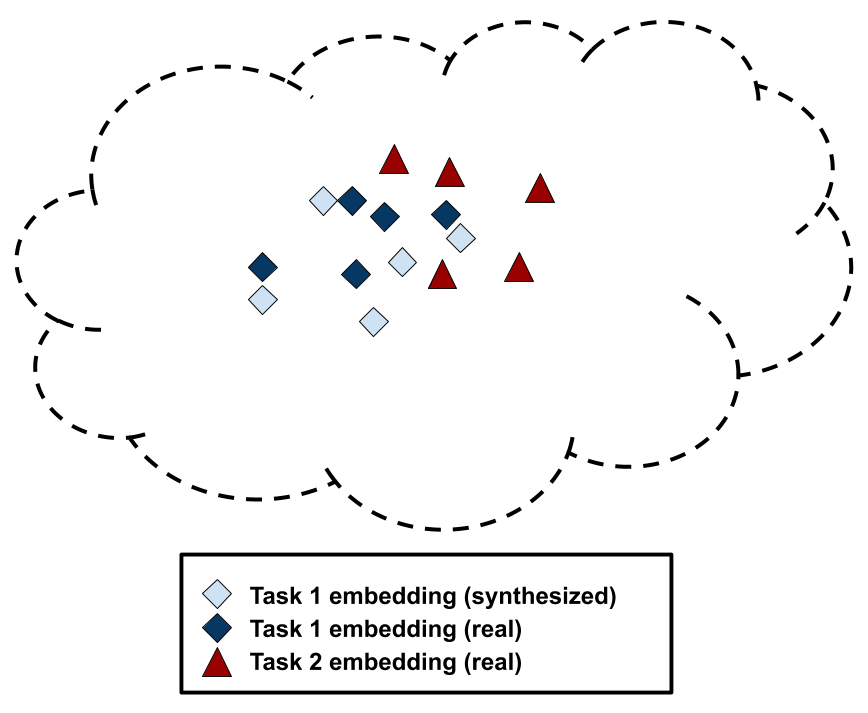}
        \caption{}
    \end{subfigure}
    \hfill
    \begin{subfigure}[t]{0.32\textwidth}
        \centering
        
        \includegraphics[width=1\textwidth]{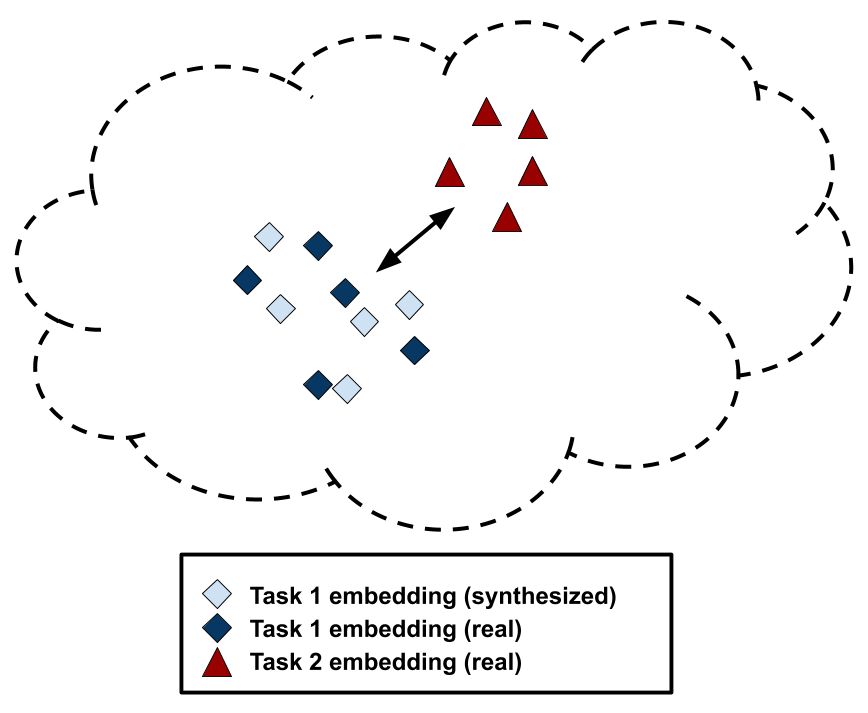}
        \caption{}
    \end{subfigure}
    \vspace{1mm}
    \caption{
    The distribution of feature embeddings when using synthetic replay data for class-incremental learning. (a) A straight application of synthetic data makes the model learn features more distinguishable between real and fake instead of task 1 and 2. This is the main problem analyzed and addressed in this work. (b) Modifying classification loss and adding regularization  mitigates the feature drifting between real and fake. (c) This is the desired feature distributions. Our method makes task 1 and 2 more separable.
    }
    \vspace{-2mm}
    \label{fig:idea}
\end{figure*}

\section{Introduction}

A shortcoming of modern computer vision settings is that they often assume offline training with a large dataset encompassing all objects to be encountered during deployment. In practice, many applications require a model be continuously updated after new environments/situations are encountered. This is the \textit{class-incremental learning} paradigm (also known as a subset of continual or lifelong learning), with the loss of knowledge over sequences of learning tasks referred to as \textit{catastrophic forgetting}. Successful incremental learning approaches have an unfortunate commonality: they require extensive memory for replay of previously seen or modeled data to avoid the catastrophic forgetting problem. This is concerning for many computer vision applications because 1) Many computer vision applications are on-device and therefore memory constrained~\citep{guo2018cloud,liu2020lifelong,soma2020fast}, and 2) Many computer vision applications learn from data which cannot be legally stored~\citep{beaulieu2019privacy,das2017assisting,Zhu_2020_CVPR}. This leads us to ask: \textit{How can computer vision systems incrementally incorporate new information without storing data?} We refer to this setting as \textit{Data-Free Class-Incremental Learning (DFCIL)} (also known as Data-Free Continual Learning~\citep{yin2020dreaming}).

An intuitive approach for DFCIL is to simultaneously train a generative model to be sampled for replay~\citep{kamra2017deep,kemker2018fear,Shin:2017,wang2020triple}. Unfortunately, training a generative model is much more computationally and memory intensive compared to a classification model. Additionally, it is not clear whether generating images from the data distribution will violate data legality concerns because using a  generative model increases the chance of memorizing potentially sensitive data~\cite{nagarajan2018theoretical}. Instead, we explore the concept of \textit{model-inversion image synthesis}, where we can invert the already provided inference network to obtain images with similar activations in the network to the training data. This idea is inviting because it requires no additional networks to be trained (it only requires the existing inference network) and is less susceptible to data privacy concerns.

The closest existing work for the DFCIL problem is DeepInversion~\citep{yin2020dreaming}, which optimizes random noise into images for knowledge distillation using a frozen teacher network. DeepInversion is designed for standard student-teacher knowledge distillation and achieves state-of-the-art performance for this task. Unfortunately, the authors report that when trying class-incremental learning for tasks where the old images and new images are similar (such as tasks from the same dataset, a standard benchmarking practice for class-incremental learning), their method performs ``statistically equivalent or slightly worse compared to Learning without Forgetting (LwF)", with LwF~\citep{li2016learning} being their most competitive existing baseline.

The goal of this paper (summarized in Figure~\ref{fig:idea}) is to dissect the cause of this failure and propose a solution for DFCIL. Specifically, we reason that when training a model with real images from the current task and synthetic images representing the past tasks, the feature extraction model causes the feature distributions of \textbf{real images from the past tasks} (which are not available during training) to be \textit{close} in the feature space to the \textbf{real images from the current task} and \textit{far} in the feature space from the \textbf{synthetic images}. This causes a bias for the model to falsely predict real images from the previous tasks with current task labels. This phenomena indicates that when training a network with two distributions of data, containing both a semantic shift (past tasks versus current task) and a distribution shift (synthetic data versus real data), the distribution shift has a higher effect on the feature embeddings. Thus, validation/test images from the previous classes will be identified as new classes due to the model fixating on their domain (i.e., realistic versus synthetic pixel distribution) rather than their semantic content (i.e., past versus current task).

To address this issue, we propose a novel class-incremental learning method which learns features for the new task with a \textit{local} classification loss which excludes the synthetic data and past-task linear heads, instead relying on importance-weighted feature distillation and linear head fine-tuning to separate feature embeddings of the new and past tasks. We show that our method represents the new state of the art for the DFCIL setting, resulting in up to a 25.1\% increase in final task accuracy (absolute difference) compared to DeepInversion for common class-incremental benchmarks, and even outperforms popular replay baselines Naive Rehearsal and LwF with a coreset. \textit{In summary, we make the following contributions}:
\begin{enumerate}[topsep=0pt,itemsep=-1ex,partopsep=1ex,parsep=1ex,leftmargin=*,labelindent=0pt]
    \itemsep 0mm
    
    \item We use a classic class-incremental learning benchmark to diagnose and analyze why standard distillation approaches for class-incremental learning (such as DeepInversion) fail when using synthetic replay data.
    
    \item We directly address this failure with a modified cross-entropy minimization, importance-weighted feature distillation, and linear head fine-tuning.
    
    \item We achieve a new state of the art performance for the DFCIL setting.
\end{enumerate}

\section{Background and Related Work}

\noindent
\textbf{Catastrophic Forgetting}: 
Approaches to mitigate catastrophic forgetting can be organized into a few broad categories and are all useful depending on which constraints are present. For example, methods which expand a model's architecture as new tasks are encountered are highly effective for applications where a model growing with tasks is practical~~\citep{ebrahimi2020adversarial,lee2020neural,lomonaco2017core50,maltoni2019continuous,Rusu:2016,smith2021unsupervised}. We do not consider these methods because the model parameters grow linearly with the number of tasks. Experience replay with stored data~\citep{aljundi2019online, aljundi2019gradient, chaudhry2018efficient,chaudhry2019episodic,Gepperth:2017,hayes2018memory,hou2019learning,Kemker:2017,Lopez-Paz:2017,Rebuffi:2016,robins1995catastrophic, rolnick2019experience,von2019continual} or samples from a generative model~\citep{kamra2017deep,kemker2018fear,ostapenko2019learning,Shin:2017,van2020brain} is highly effective when storing training data or training/saving a generative model is possible, unlike the DFCIL setting. 

Another approach is to regularize the model with respect to past task knowledge while training the new task. This can either be done by regularizing the model in the  weight space (i.e., penalize changes to model parameters)~\citep{aljundi2017memory,ebrahimi2019uncertainty,kirkpatrick2017overcoming,titsias2019functional,zenke2017continual} or the prediction space (i.e., penalize changes to model predictions)~\citep{castro2018end,hou2018lifelong,lee2019overcoming,li2016learning,smith2021memory}. Prediction space regularization (accomplished using \textit{knowledge distillation}) has been found to perform better than model regularization based methods for class-incremental learning~\citep{lesort2019generative,van2018generative}.

\noindent
\textbf{Knowledge Distillation in Class-Incremental Learning}:  Based on the original work proposing knowledge distillation from a large model to a smaller model~\citep{hinton2015distilling}, methods such as Learning Without Forgetting~\citep{li2016learning}, Distillation and Retrospection~\citep{hou2018lifelong}, End-To-End Incremental Learning~\citep{castro2018end}, Global Distillation~\cite{lee2019overcoming}, and Bias Correction~\citep{wu2019large} have effectively leveraged knowledge distillation as a prediction regularization technique for incremental learning.  The high level idea of knowledge distillation is to periodically save a frozen copy of the model (here we use a ConvNet) and to ensure the new model makes similar predictions to the frozen model over a set of distillation images (while simultaneously learning the new task). Knowledge distillation does not require that the frozen model be replaced at the task sequence boundaries, but this is typically done when evaluating competing methods. This regularization can also take place in the feature space rather than the prediction space~\citep{balaji2020effectiveness,hou2019learning}, which we refer to as \textit{feature distillation}. These knowledge distillation methods require stored data over which to enforce similar predictions, but the next section describes a form of knowledge distillation which \text{does not require training data}.

\noindent
\textbf{Data-Free Knowledge Distillation}: Knowledge from a neural network can be transferred in the absence of training data. We refer to the line of work which \textit{synthesizes} distillation images using the trained inference network itself and resulting activation statistics as \textit{data-free knowledge distillation}. This approach is very important for applications where training data is sensitive and not easily available  for legality issues. The first such work we are aware of, DeepDream~\citep{mordvintsev2015inceptionism}, optimizes randomly generated noise into images which minimize classification loss and an image prior. Another early method~\citep{lopes2017data} matches stored layer statistics from a trained ``teacher" model while leaving a small memory footprint using frequency-based compression techniques. The Data-Free Learning method~\citep{chen2019data} exploits a GAN architecture to synthesize images which match the trained teacher's statistics while balancing content losses which maximize both temperature scaled linear heads (to drive class-specific content high) and class prediction entropy (to encourage high diversity of classes sampled). Three recent methods leverage layer content stored in batch-normalization layers to synthesize realistic looking images for knowledge distillation~\citep{haroush2020knowledge,luo2020large,yin2020dreaming}. 

To the best of our knowledge, only two class-incremental learning methods are designed for the data-free perspective. Automatic-Recall Machines (ARM)~\citep{ji2020automatic} perturb training data from the current task into images which maximize ``forgetting" from the past tasks. However, this method is designed for a ``single pass" setting where data is only trained for one epoch, which is a different setting than ours. 
DeepInversion~\citep{yin2020dreaming} also evaluates  data-free knowledge distillation in the class-incremental learning paradigm, but only found success in small task sequences (max of three) using tasks which are very distinct in image content. Our paper dissects why the DeepInversion method fails at difficult class-incremental learning problems and proposes a solution for successful data-free class-incremental learning.

\section{Preliminaries}

\noindent
\textbf{Class-Incremental Learning}: In class-incremental learning, a model is shown labeled data corresponding to $M$ semantic object classes $c_1, c_2, \dots, c_M$ over a series of $N$ tasks corresponding to non-overlapping subsets of classes. We use the notation $\mathcal{T}_n$ to denote the set of classes introduced in task $n$, with $|\mathcal{T}_n|$ denoting the number of object classes in task $n$. Each class appears in only a single task, and the goal is to incrementally learn to classify new object classes as they are introduced while retaining performance on previously learned classes. The class-incremental learning setting is challenging because no task indexes are provided to the learner during inference and the learner must support classification across all classes seen up to task $n$~\cite{hsu2018re}. This is more difficult than \textit{task-incremental learning}, where the task indexes are given during both training and inference. While our setting does not necessitate known task boundaries during training, we follow prior works~\citep{li2016learning,hsu2018re,yin2020dreaming} for fair comparison and create model copies at the task boundaries for each method.

To describe our inference model, we denote $\theta_{i,n}$ as the model $\theta$ at time $i$ that has been trained with the classes from task $n$. For example, $\theta_{n,1:n}$ refers to the model trained during task $n$ and its logits associated with all tasks up to and including class $n$. We drop the second index when describing the model trained during task $n$ with all logits (for example, $\theta_{n}$).

\section{Baseline Approach for Data-Free Class Incremental Learning}

In this section, we propose a general baseline for data-free class incremental learning based on efforts in prior work. We start by summarizing the \textit{data synthesis} (i.e. generation of images from the discriminative model itself) approach we found most successful for class-incremental learning. We then review pertinent \textit{knowledge distillation} losses, ultimately focusing on the loss functions used by DeepInversion~\citep{yin2020dreaming} for class-incremental learning.

\noindent
\textbf{Model-Inversion Image Synthesis}:
Most model-inversion image synthesis approaches seek to synthesize images by directly optimizing them with respect to a prior discriminative model $\theta_{n-1}$.  For $B$ synthetic images, a tensor $\hat{X} \in \mathcal{R}^{B \times H \times W \times C} = \{ \hat{x}_1 \cdots \hat{x}_b \}$, where $H$, $W$, and $C$ correspond to the training data image dimensions, is initialized from Gaussian noise. However, it is computationally inefficient to optimize one batch of images at a time. Especially considering that class-incremental learning is expected to be computationally efficient, we choose to approximate this optimization from noise to synthesized images using a ConvNet-parameterized function $\mathcal{F}_{\phi}$. This allows the framework to train $\mathcal{F}_{\phi}$ once per task (using \textit{only} $\theta_{n-1}$, i.e., no data), only store it temporarily during that given task, sample synthetic images as needed during training for task $\mathcal{T}_{n}$, and then discarded it at the end of the task. 

$\mathcal{F}_{\phi}$ can struggle with synthetic class diversity; rather than condition $\mathcal{F}_{\phi}$ on class labels $Y$, we follow~\citep{chen2019data} and optimize the diversity of class predictions of synthetic images to match a uniform distribution. Denoting $p_{\theta}(x)$ as the predicted class distribution produced by model $\theta$ for some input $x$, we want to \textit{maximize} the entropy of the mean class prediction vector for synthetic samples $\hat{X}$. Formally, we minimize a \textit{label diversity loss}:
\begin{equation}
    \mathcal{L}_{div}(\hat{Y}) = -\mathcal{H}_{info} \left( \frac{1}{B} \sum_b p_{\theta}(\hat{x}_b) \right)
\end{equation}
where $\mathcal{H}_{info}$ is the information entropy. Notice when the loss is taken at the minimum, every element in the mean class prediction vector would equal $\frac{1}{|\mathcal{T}_{1} \dots \mathcal{T}_{n-1}|}$, meaning that classes are generated with at roughly the same rate.

In addition to diversity, to consistently synthesize useful images in the DFCIL setting, the images must enforce calibrated class confidences, consistency of feature statistics, and a locally smooth latent space, described below.

\textit{Content loss}, $\mathcal{L}_{con}$,  maximizes class prediction confidence with respect to the image tensor $\hat{X}$ such that $\theta_{n-1}$ should make confident predictions on all inputs. Formally, $\mathcal{L}_{con}$ is the cross entropy classification loss between the class predictions of $\hat{X}$ and the maximum class predictions $\hat{Y}$:
\begin{equation}
    \mathcal{L}_{con}(\hat{X}, \hat{Y}) = \mathcal{L}_{CE} \left( p^{\alpha_{temp}}_{\theta_{n-1,1:n-1}}(\hat{x}), \hat{y} \right)
\end{equation}
\begin{equation}
    \hat{y} = \underset{\hat{y} \in \mathcal{T}_{1} \dots \mathcal{T}_{n-1}}{argmax} \hspace{0.5em} p_{\theta_{n-1, 1:n-1}}(\hat{x})
\end{equation}
where $\mathcal{L}_{CE}$ is standard cross entropy loss and the logit output of $\theta$ is scaled by a temperature constant $\alpha_{temp}$. 
By combining $\mathcal{L}_{con}$ with $\mathcal{L}_{div}$, we ensure that the synthesized images will represent the distribution of all past task classes

Prior work has found that the complexity of model-inversion can cause the distribution of $\theta_{n-1}$ features to deviate greatly from distributions over batches of synthetic images. Intuitively, the batch statistics of synthesized images should match those of the batch normalization layers in $\theta_{n-1}$. To enforce this, \textit{stat alignment loss}, $\mathcal{L}_{stat}$, penalizes the deviation between intermediate layer  batch-normalization statistics (BNS) stored in $\theta_{n-1}$ and features at those layers for synthetic images~\citep{haroush2020knowledge,luo2020large,yin2020dreaming}:
\begin{equation}
    \mathcal{L}_{stat}(\hat{X}) = \frac{1}{L} \sum_{l=1}^L BNS(\hat{\mu}_{\hat{X},l},\hat{\sigma}_{\hat{X},l},\mu_l,\sigma_l)
\end{equation}
\begin{equation}
\begin{split}
    BNS&(\hat{\mu},\hat{\sigma},\mu,\sigma) = KL(\mathcal{N}(\mu,\sigma^2)) || \mathcal{N}(\hat{\mu},\hat{\sigma}^2) \\
    &= \log{\frac{\hat{\sigma}}{\sigma}} - \frac{1}{2} \left( 1 - \frac{\sigma^2 + (\mu - \hat{\mu})^2}{\hat{\sigma}^2} \right)
\end{split}
\end{equation}
where $KL$ denotes the Kullback-Leibler (KL) divergence, $\hat{\mu}_{\hat{X},l},\hat{\sigma}_{\hat{X},l}$ are the mean and standard deviation of the features at layer $l$ for a given mini-batch of synthesized image, and $\mu_l,\sigma_l$ are the batch-norm statistics of said layer $l$. Because batch statistics of $\theta_{n-1}$ are stored in batch normalization layers, this loss does not require any additional storage.

Additionally, prior knowledge tells us natural images are more locally smooth in pixel space than the initial noise. As such, we can stabilize the optimization by minimizing the \textit{smoothness prior loss} $\mathcal{L}_{prior}$. Formally, $\mathcal{L}_{prior}$ is the L2 distance between each synthesized image ($\hat{x}$) and a version blurred with Gaussian kernel ($\hat{x}_{blur}$):
\begin{equation}
    \mathcal{L}_{prior}(\hat{X}) = || \hat{x} - \hat{x}_{blur} ||^2_2
\end{equation}

All together, assuming the use of $\mathcal{F}_{\phi}$ for efficiency, the final loss for the baseline is therefore:
\begin{equation}
\begin{split}
    \underset{\mathcal{F}_{\phi}}{min} & \hspace{0.5em} \alpha_{con} \mathcal{L}_{con}(\hat{X},\hat{Y} ) + \alpha_{div}  \mathcal{L}_{div}(\hat{Y}  ) \\
    &+ \alpha_{stat} \mathcal{L}_{stat}(\hat{X} ) + \alpha_{prior} \mathcal{L}_{prior}(\hat{X} )
\end{split}
\label{eq:inversion}
\end{equation}
Importantly: although we optimize $\mathcal{F}_{\phi}$ rather than $\hat{X}$ as done in~\citep{haroush2020knowledge,yin2020dreaming}, this method can use the latter with a sacrifice to computational efficiency.
\begin{figure*}[t]
    \centering
    \begin{subfigure}[t]{0.42\textwidth}
        \centering
        \includegraphics[width=1\textwidth, trim = {0 0 0 0}, clip]{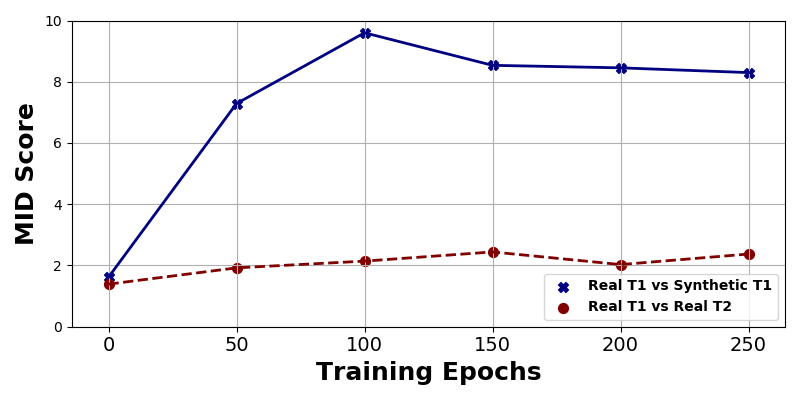}
        \caption{DeepInversion~\citep{yin2020dreaming}}
    \end{subfigure}
    \hspace{.5cm}
    \begin{subfigure}[t]{0.42\textwidth}
        \centering
        \includegraphics[width=1\textwidth, trim = {0 0 0 0}, clip]{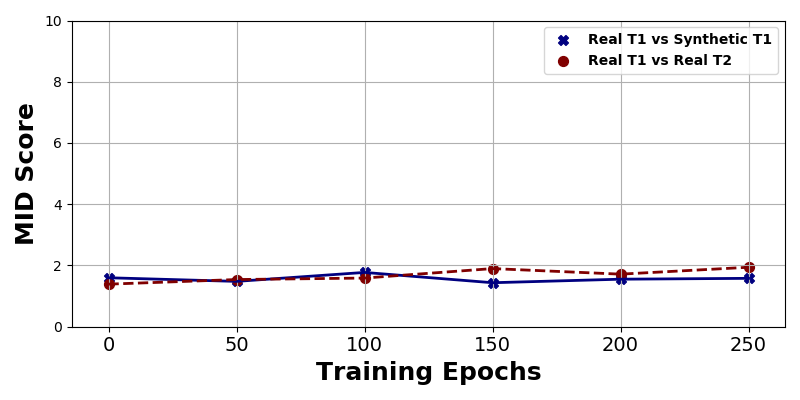}
        \caption{Our Method}
    \end{subfigure}
    \caption{
    Representational distance scores (MID) between feature embeddings of real task 1 data and synthetic task 1 data (blue), real task 2 data (red). Task 1 corresponds to ten classes of CIFAR-100 while task 2 corresponds to a different ten classes of CIFAR-100; the results are generated after training on task 2.
    }
    \label{fig:cka}
\end{figure*}

\vspace{0.2cm}
\noindent
\textbf{Distilling Synthetic Data for Class-Incremental Learning}:
In the class-incremental learning setting, where the classes of task $\mathcal{T}_n$ are modeled without unlearning the representation of classes of tasks $\mathcal{T}_1 \cdots \mathcal{T}_{n-1}$, knowledge distillation over the synthesized images is most often used to regularizes $\theta_n$, forcing it to learn $\mathcal{T}_n$ with minimal degradation to $\mathcal{T}_1 \cdots \mathcal{T}_{n-1}$ knowledge. For task $\mathcal{T}_n$, we synthesize images from a frozen copy of ($\theta_{n-1}$) trained during task $\mathcal{T}_{n-1}$. These synthetic images then help us distill knowledge learned in tasks $\mathcal{T}_1 \cdots \mathcal{T}_{n-1}$ into our current model ($\theta_n$) as it learns from $\mathcal{T}_n$ data. 

In our baseline, we adopt the distillation approach used in DeepInversion~\citep{yin2020dreaming}, which generalizes the original Learning without Forgetting (LwF)~\citep{li2016learning} distillation approach. Formally, given current task data, $x$, and synthesized distillation data, $\hat{x}$, we minimize:
\begin{equation}
\begin{split}
    \underset{\theta_n}{min} \hspace{0.5em} \mathcal{L}_{CE} \left( p_{\theta_{n,1:n}}(x), y \right) &+ \mathcal{L}_{KD}^{DI}(x,\theta_n,\theta_{n-1}) \\
    &+ \mathcal{L}_{KD}^{DI}(\hat{x},\theta_n,\theta_{n-1})
\end{split}
\label{eq:lwf-b}
\end{equation}
where $\mathcal{L}_{KD}^{DI}$ is a knowledge distillation regularization as:
\begin{equation}
    \mathcal{L}_{KD}^{DI}(x,\theta_n,\theta_{n-1}) = KL(p_{\theta_{n-1,1:n}}(x) || p_{\theta_{n,1:n}}(x)) 
    \label{eq:kd-di}
\end{equation}

Here, $p_{\theta_{n-1,1:n}}(x)$ is simply $p_{\theta_{n-1,1:n-1}}(x)$ appended with zeros to represent zero class-probability for the new classes (which are not available for $\theta_{n-1,1:n-1}$). The key idea of $\mathcal{L}_{KD}^{DI}$ is that, as the logits of the teacher and student will always be different in the class incremental learning setting, appending zeros to the class-probability vectors aligns the student and teacher logit dimensions for a better transfer of knowledge.
\section{Diagnosis: Feature Embedding Prioritizes Domains Over Semantics}

To understand why the baseline approach for DFCIL fails, we analyze \textit{representational distance} between embedded features with a metric that captures the distance between mean embedded images of two distribution samples. Specifically, we assign a Mean Image Distance (MID) score between a reference sample of images $x_a$ and another sample of images $x_b$, where a \textit{high score} indicates \textit{dissimilar} features and a \textit{low score} indicates \textit{similar} features. We calculate this score as:
\begin{equation}
    MID(z_a,z_b) = \left\Vert \frac{\mu_a - \mu_b}{\sigma_a} \right\Vert_2
\end{equation}
where $z_a, z_b$ is the penultimate feature embedding of $x_a, x_b$; $\mu_a, \mu_b$ are the mean image feature embeddings of $x_a, x_b$; and $\sigma_a^2$ is the feature variance of $x_a$. We normalize the distance between mean embedded images by the standard deviation of the reference distribution sample $x_a$ to minimize the impact of highly deviating features. Additional analysis using Maximum Mean Discrepancy (MMD)~\citep{gretton2012kernel} is available in our Appendix (\ref{app:mmd}).

For our analysis, we start by training our model for the first two tasks in the ten-task CIFAR-100 benchmark described in Section~\ref{sec:experiments}. We calculate MID between feature embeddings of real task 1 data and real task 2 data, and then we calculate MID between feature embeddings of real task 1 data and synthetic task 1 data. The results are reported in Figure~\ref{fig:cka}. For (a) DeepInversion, the MID score between real task 1 data and synthetic task 1 data is significantly higher than the MID score between real task 1 data and real task 2 data. This indicates that the embedding space prioritizes \textit{domain} over \textit{semantics}, which is detrimental because the classifier will learn the decision boundary between synthetic task 1 and real task 2, introducing great classification error with real task 1 images. This diagnosis motivates our approach, which is proposed in the following section. For (b) our method, the MID score between real task 1 data and synthetic task 1 data is much lower, indicating that our feature embedding prioritizes \textit{semantics} over \textit{domain}.

\section{A New Distillation Strategy for DFCIL}

\begin{figure*}[t!]
    \centering
    \includegraphics[height=130pt]{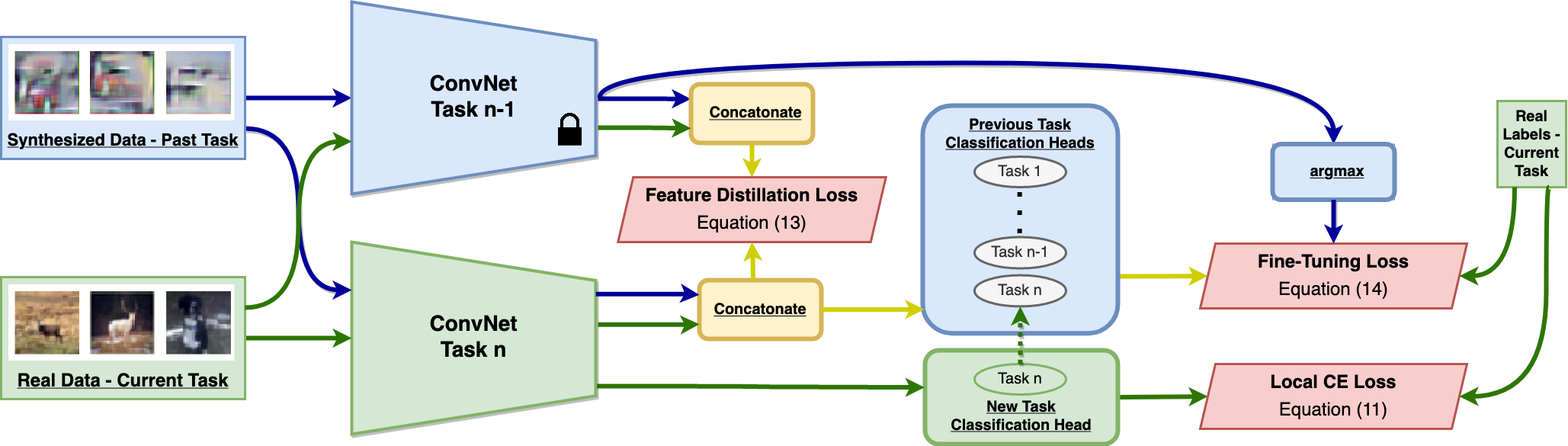}
    \vspace{0.5cm}
	\caption{Our approach combines (i) learning features for the new task with Eq.~\eqref{eq:ce}, (ii) minimizing feature drift over the previous task with Eq.~\eqref{eq:kd-hard}, and (iii) separating class overlap between new and previous classes in the embedding space with Eq.~\eqref{eq:ft}. We use blue arrows to designate the compute path of the synthetic previous tasks data, green arrows to designate the compute path of the real current task data, and yellow arrows to designated the compute path of both real and synthetic data. We separate out the task $\mathcal{T}_n$ head to show that the local CE loss Eq.~\eqref{eq:ce} uses only this head.}
	\label{fig-approach}
	\vspace{-4mm}
\end{figure*}

We take the perspective that continual learning should balance: (i) learning features for the new task, (ii) minimizing feature drift over the previous task, and (iii) separating class overlap between new and previous classes in the embedding space (this is similarly discussed in another work~\citep{hou2019learning} under a different setting). Generally, (i) and (iii) are simultaneously achieved with $\mathcal{L}_{CE}$, but we argue that by separating this into two different losses, features for the new task are learned which do not discriminate between real and synthetic images (i.e. avoid the feature domain bias problem). Following this idea, we propose a new class-incremental learning approach designed for DFCIL which addresses each of these goals independently, as described in the rest of this section.

\noindent
\textbf{Learning current task features}: The intuition behind our method is to learn features for our current task while circumventing the feature embedding for real data becoming highly biased towards the most recent task. That is, we form $\mathcal{L}_{CE}$ so that the likelihood of $x$ being real versus synthetic, $p_{\theta_{n,n }}(x \in X_{real})$, is not helpful in its minimization. We do this by computing cross entropy classification loss \textit{locally} across the new class linear heads without including the past class linear heads. With this formation, we prevent the model from learning to separate the new and past class data via domain (i.e. synthetic vs. real). Formally, we minimize:
\begin{equation}
\begin{split}
    \mathcal{L}_{CE}^{\mathcal{T}_n} \left( p_{\theta_{n,n}}(x), y \right) = \mathcal{L}_{CE} \left( p_{\theta_{n,n }}(x|y \in \mathcal{T}_n), y \right)
\end{split}
\label{eq:ce}
\end{equation}

\noindent
\textbf{Minimizing feature drift over previous task data}: 
Because our distillation images are of another domain than the real current task images (causing the feature domain bias problem), we seek a loss function which directly alleviates forgetting in the feature space. An alternative to standard knowledge distillation over softmax predictions ($\mathcal{L}_{KD}^{DI}$) is \textit{feature distillation}, which instead distils the feature content from the penultimate layer. This is formally given as:
\begin{equation}
    \mathcal{L}_{KD}^{feat}(\cdot) = ||\theta_{n,1:n-1}^{L-1}(x) - \theta_{n-1,1:n-1}^{L-1}(x) ||^2_2
    \label{eq:kd-feature}
\end{equation}
where $L-1$ denotes the penultimate layer output of the model. Our intuition is that there exists a trade-off between standard knowledge distillation and feature distillation. $\mathcal{L}_{KD}^{feat}$ reinforces important components of past task data, but it is a strong regularization which inhibits the plasticity of the model (and its ability to learn the new task). On the other hand, $\mathcal{L}_{KD}^{DI}$ does not inhibit learning the new task but can be minimized with a solution in which feature drift has occurred, resulting in the real vs synthetic bias. 

Instead, we desire an \textit{importance-weighted} feature distillation which reinforces only the most important components of past task data while allowing less important features to be adapted for the new task. We simply use the linear heads of $\mathcal{T}_1 \cdots \mathcal{T}_{n-1}$ from the frozen model $\theta_{n-1}$, or:
\[
    \mathcal{L}_{KD}^{wfeat}(\cdot) = || \mathcal{W}\left( \theta_{n,1:n-1}^{L-1}(x) \right) - \mathcal{W}\left(\theta_{n-1,1:n-1}^{L-1}(x)\right) ||^2_2
\]
\begin{equation}
    \text{where} \quad \mathcal{W} = \theta_{n-1,1:n-1}^{L}
\label{eq:kd-hard}
\end{equation}
By using this importance-weight matrix, features associated with a high magnitude in $\mathcal{W}$ are more important to preserve. In this way, the frozen linear heads from the past tasks indicate approximately how much a change to each feature affects the class distribution.

\noindent
\textbf{Separating Current and Past Decision Boundaries}: Finally, we need to separate the decision boundaries of the current and past classes without allowing the feature space to distinguish between the real and synthetic data. We do this by fine-tuning the linear classification head of $\theta_{n,1:n}$ with standard cross-entropy loss. Importantly, this loss does not update any parameters in $\theta_{n,1:n}$ besides the final linear classification head. Formally, we minimize:
\begin{equation}
    \mathcal{L}_{FT}^{\mathcal{T}_{1:n}}\left( p_{\theta_{n,1:n}}(x,y) \right) = 
    \mathcal{L}_{CE} \left( p^*_{\theta_{n,1:n}}(x,y)\right)
    \label{eq:ft}
\end{equation}
where $p^*$ is calculated with $\theta_{n,1:n}^{1:L-1}$ (i.e., every layer in the model except the classification layer) frozen, updating only $\theta_{n,1:n}^{L}$. As done in~\citep{lee2019overcoming, Shin:2017,wu2019large}, we add task-balancing loss weighting to balance the contributions from the current task with the past tasks.

\noindent
\textbf{Final Objective: }
Visualized in Figure~\ref{fig-approach}, our final optimization objective is given as:
\begin{equation}
\begin{split}
    \underset{\theta_n}{min} \hspace{0.5em} & \mathcal{L}_{CE}^{\mathcal{T}_n} \left( p_{\theta_{n,n}}(x), y \right) + \lambda_{kd} \mathcal{L}_{KD}^{wfeat}(\{ x, \hat{x}\},\theta_n,\theta_{n-1}) \\
    &+ \lambda_{ft} \mathcal{L}_{FT}^{\mathcal{T}_{1:n}}\left( p_{\theta_{n,1:n}}(\{ x, \hat{x}\}), \{y, \hat{y} \} \right)
\end{split}
\label{eq:ours}
\end{equation}
where the $\lambda$ terms weight the contributions of $\mathcal{L}_{KD}^{wfeat}$ and $\mathcal{L}_{FT}$ with respect to $\mathcal{L}_{CE}^{\mathcal{T}_n}$

\begin{table*}[t]
\small
\caption{Results (\%) for \textit{data-free} class-incremental learning on CIFAR-100 for various numbers of tasks (5, 10, 20). Results are reported as an average of 3 runs.}
\centering
\begin{tabular}{c | c | c c | c c | c c}
    \hline 
    \multicolumn{2}{c|}{Tasks} & \multicolumn{2}{c|}{5} & \multicolumn{2}{c|}{10} & \multicolumn{2}{c}{20} \\
    \hline
    Method & Replay Data & $A_{N}$ ($\uparrow$)  & $\Omega$ ($\uparrow$) & $A_{N}$ ($\uparrow$)  & $\Omega$ ($\uparrow$) & $A_{N}$ ($\uparrow$)  & $\Omega$ ($\uparrow$)\\
    \hline
    Upper Bound & None & $ 69.9 \pm 0.2 $ & $ 100.0 \pm 0.0 $ & $ 69.9 \pm 0.2 $ & $ 100.0 \pm 0.0 $ & $ 69.9 \pm 0.2 $ & $ 100.0 \pm 0.0 $ \\
    \hline
    Base & None & $ 16.4 \pm 0.4 $ & $ 48.9 \pm 1.1 $ & $ 8.8 \pm 0.1 $ & $ 32.1 \pm 1.1 $ & $ 4.4 \pm 0.3 $ & $ 19.7 \pm 0.7 $  \\ 
    DGR~\citep{Shin:2017} & Generator & $ 14.4 \pm 0.4 $ & $ 45.5 \pm 0.9 $ & $ 8.1 \pm 0.1 $ & $ 30.5 \pm 0.6 $ & $ 4.1 \pm 0.3 $ & $ 19.0 \pm 0.3 $  \\
    LwF~\citep{li2016learning} & None & $ 17.0 \pm 0.1 $ & $ 49.5 \pm 0.1 $  & $ 9.2 \pm 0.0 $ & $ 33.3 \pm 0.9 $ & $ 4.7 \pm 0.1 $ & $ 20.1 \pm 0.3 $  \\ 
    LwF~\citep{li2016learning} & Synthetic & $ 16.7 \pm 0.1 $ & $ 49.8 \pm 0.1 $ & $ 8.9 \pm 0.0 $ & $ 32.3 \pm 0.0 $ & $ 4.7 \pm 0.0 $ & $ 19.7 \pm 0.0 $  \\
    DeepInversion~\citep{yin2020dreaming} & Synthetic & $ 18.8 \pm 0.3 $ & $ 53.2 \pm 0.9 $ & $ 10.9 \pm 0.6 $ & $ 37.9 \pm 0.8 $ & $ 5.7 \pm 0.3 $ & $ 23.6 \pm 0.7 $  \\ 
    \hline 
    Ours & Synthetic & $ \mathbf{43.9 \pm 0.9 } $ & $ \mathbf{ 78.6 \pm 1.1 } $ & $\mathbf{ 33.7 \pm 1.2 } $ & $ \mathbf{ 69.6 \pm 1.6 } $ & $ \mathbf{ 20.0 \pm 1.4 } $ & $ \mathbf{ 52.5 \pm 2.5 } $ \\
    \hline
\end{tabular}
\vspace{-3mm}
\label{tab:results_cifar100_datafree}
\end{table*}

\begin{table*}[t]
\small
\caption{Results (\%) for class-incremental learning \textit{with replay data} on CIFAR-100 for various numbers of tasks (5, 10, 20). A coreset of 2000 images is leveraged for replay-based methods, and thus \textit{these methods do not meet problem the DFCIL constraints} (note we report for our method numbers \textit{without} any coreset). Results are reported as an average of 3 runs.}
\centering
\begin{tabular}{c | c | c c | c c | c c}
    \hline 
    \multicolumn{2}{c|}{Tasks} & \multicolumn{2}{c|}{5} & \multicolumn{2}{c|}{10} & \multicolumn{2}{c}{20} \\
    \hline
    Method & Replay Data & $A_{N}$ ($\uparrow$)  & $\Omega$ ($\uparrow$) & $A_{N}$ ($\uparrow$)  & $\Omega$ ($\uparrow$) & $A_{N}$ ($\uparrow$)  & $\Omega$ ($\uparrow$)\\
    \hline
    Upper Bound & None & $ 69.9 \pm 0.2 $ & $ 100.0 \pm 0.0 $ & $ 69.9 \pm 0.2 $ & $ 100.0 \pm 0.0 $ & $ 69.9 \pm 0.2 $ & $ 100.0 \pm 0.0 $ \\
    \hline
    Naive Rehearsal & Coreset & $ 34.0 \pm 0.2 $ & $ 73.4 \pm 0.8 $ & $ 24.0 \pm 1.0 $ & $ 64.6 \pm 2.1 $ & $ 14.9 \pm 0.7 $ & $ 51.4 \pm 2.9 $  \\ 
    LwF~\citep{li2016learning} & Coreset& $ 39.4 \pm 0.3 $ & $ 79.0 \pm 0.0  $ & $ 27.4 \pm 0.8 $ & $ 69.4 \pm 0.4 $ & $ 16.6 \pm 0.4 $ & $ 54.2 \pm 2.2 $  \\ 
    BiC~\citep{wu2019large} & Coreset & $ \mathbf{53.7 \pm 0.4} $ & $ \mathbf{87.5 \pm 0.9} $ & $ \mathbf{45.9 \pm 1.8} $ & $ \mathbf{81.9 \pm 2.0} $ & $ \mathbf{37.5 \pm 3.2} $ & $ \mathbf{71.7 \pm 3.4} $  \\
    \hline 
    Ours & Synthetic & $ 43.9 \pm 0.9  $ & $ 78.6 \pm 1.1  $ & $ 33.7 \pm 1.2  $ & $  69.6 \pm 1.6  $ & $  20.0 \pm 1.4  $ & $ 52.5 \pm 2.5 $ \\
    \hline
\end{tabular}
\vspace{-3mm}
\label{tab:results_cifar100_replay}
\end{table*}

\begin{table*}[t]
\small
\caption{Ablation Study Results (\%) for ten-task class-incremental learning on CIFAR-100. Results are reported as an average of 3 runs.}
\centering
\begin{tabular}{c | c c}
    \hline
    Metric ($\uparrow$) & $A_{N}$  & $\Omega$ \\
    \hline
    Full Method  & $\mathbf{33.7 \pm 1.2} $ & $\mathbf{69.6 \pm 1.6} $ \\ 
    \hline
    Ablate Task Balancing Loss Weighting~\citep{lee2019overcoming, Shin:2017,wu2019large} & $ 23.4 \pm 1.5 $ & $ 62.2 \pm 2.4 $  \\
    Replace Modified CE Loss, Eq.~\eqref{eq:ce}, w/ Standard CE Loss & $ 16.5 \pm 0.5 $ & $ 46.6 \pm 0.9 $  \\ 
    Ablate Real Data Distillation: Eq.~\eqref{eq:kd-hard} w/ $X$ & $ 15.9 \pm 2.1 $ & $ 58.9 \pm 3.2 $  \\ 
    Ablate Synthetic Data Distillation: Eq.~\eqref{eq:kd-hard} w/ $\hat{X}$ & $ 12.7 \pm 7.4 $ & $ 55.3 \pm 8.1 $  \\ 
    Ablate FT-CE Loss: Eq.~\eqref{eq:ft} & $ 9.8 \pm 0.6 $ & $ 35.9 \pm 1.3 $  \\ 
    \hline
\end{tabular}
\label{tab:ablation-cifar100}
\vspace{-4mm}
\end{table*}
\noindent
\section{Experiments}
\label{sec:experiments}

We evaluate our approach with several image datasets in the class incremental continual learning setting. We implemented baselines which do not store training data for rehearsal: Deep Generative Replay (DGR)~\citep{Shin:2017}, Learning without Forgetting (LwF)~\citep{li2016learning}, and Deep Inversion (DeepInversion)~\citep{yin2020dreaming}. Additionally, we report the upper bound performance (i.e., trained offline) and performance for a neural network trained only on classification loss using the new task training data (we refer to this as Base). We note that a downside of any generative method is that they (1) require long-term storage of a generative model and (2) may violate data legality concerns.

For a fair comparison, our implementation of DeepInversion uses the same image synthesis strategy as our method, with the difference being the distillation method. We do not tune hyperparameters on the full task set because tuning hyperparameters with hold out data from all tasks may violate the principal of continual learning that states each task is visited only once~\citep{Ven:2019}. Importantly, we shuffle the class order before sampling tasks, and do this with a consistent seed (different for each trial) so that results can be directly compared. 
We include supplementary details and metrics in our Appendix: additional results (\ref{app:full_results}), additional experiment details (\ref{app:exp_deets}), and hyperparameter selection (\ref{app:param_sweeps}).

\noindent
\textbf{Evaluation Metrics}: Following prior works, we evaluate methods in the class-incremental learning setting using: (I) final performance, or the performance with respect to all past classes after having seen all $N$ tasks (referred to as $A_{N,1:N}$); and (II) $\Omega$, or the average (over all tasks) normalized task accuracy with respect to an offline oracle method ~\citep{hayes:2019}. We use index $i$ to index tasks through time and index $n$ to index tasks with respect to test/validation data (for example, $A_{i,n}$ describes the accuracy of our model after task $i$ on task $n$ data). Specifically:
\begin{equation}
    A_{i,n} = \frac{1}{|\mathcal{D}_n^{test}|} \sum_{(x,y)\in\mathcal{D}_n^{test}} \mathbbm{1}(\hat{y}(x,\theta_{i,n}) = y \mid \hat{y} \in \mathcal{T}_n)
\end{equation}
\begin{equation} 
    \Omega = \frac{1}{N} \sum_{i=1}^N \sum_{n=1}^{i} \frac{|\mathcal{T}_{n}|}{|\mathcal{T}_{1:i}|} \frac{A_{i,1:n}}{A_{offline,1:n}}
\end{equation}
where $A_{offine}$ is the task accuracy trained in the offline setting (i.e., the upper-bound performance). $\Omega$ is designed to evaluate the global task and is therefore computed with respect to all previous classes. For the final task accuracy in our results, we will denote $A_{N,1:N}$ as simply $A_{N}$.

\noindent
\textbf{Data-Free Class-Incremental Learning - CIFAR-100 Benchmark}: Our first benchmark is ten-task class-incremental learning on the CIFAR-100 dataset~\citep{krizhevsky2009learning} which contains 100 classes of 32x32x3 images. Following prior work~\citep{wu2019large}, we train with a 32-layer ResNet~\citep{he2016deep} for 250 epochs. The learning rate is set to 0.1 and is reduced by 10 after 100, 150, and 200 epochs. We use a weight decay of 0.0002 and batch size of 128. Using a simple grid search to find hyperparameters for Eq~\eqref{eq:inversion}, we found $\{ \alpha_{con}, \alpha_{div}, \alpha_{stat}, \alpha_{prior}, \alpha_{temp} \}$ as \{1 , 1, 5e1, 1e-3, 1e3\} (these hyperparameters are not introduced by our method). For Eq.~\eqref{eq:ours} in our method, we found $\{\lambda_{kd}, \lambda_{ft}\}$ as \{1e-1 ,1\}, and we use prior reported loss-weighting hyperparameters for our implementations of other methods. We use a temperature scaling of 2 for all softmax knowledge distillation instances. We model $\mathcal{F}$ with the same parameters as~\citep{luo2020large} and train Eq.~\eqref{eq:inversion} after starting each task using 5,000 training steps of Adam optimization (learning rate 0.001).

The results are given in Table~\ref{tab:results_cifar100_datafree}. We see that our method outperforms not only the DFCIL methods (including a 25.1\% increase in final task accuracy over DeepInversion), but even the generative approach (despite their use of significant additional memory between tasks). To our surprise, we found DGR~\citep{Shin:2017} to perform poorly for class-incremental learning on this dataset (and in fact every dataset we experiment with); this finding is repeated in another recent work~\cite{van2020brain} which also found DGR to perform worse\footnote{We did not implement additional generative-replay results because that is not the focus of our paper. Instead, we compare to 1) other DFCIL methods to show our method performs best in our setting, and 2) SOTA replay-based methods to show our method performs close to SOTA despite not storing replay data.} than Base. We are not surprised to see that LwF~\citep{li2016learning} performs worse than naive rehearsal, as this is also common for class-incremental learning~\citep{van2018generative,van2020brain}. Finally, we observe that synthetic data does not improve LwF's performance. This is consistent with our finding that the feature embedding prioritizes domains over semantics when using standard distillation strategies.

\noindent
\textbf{Class-Incremental Learning with Replay Data - CIFAR-100 Benchmark}: 
In Table~\ref{tab:results_cifar100_replay}, we compare our method (which does \textit{not store replay data}) to other methods which \textit{do store replay data}. We use the same number of coreset images as the ~\citep{castro2018end,wu2019large}. We found that our method can perform \textit{significantly greater than LwF and Rehearsal, which store replay data}. We also compare our method with a SOTA replay-based class-incremental learning method: Bias Correction (BiC)~\citep{wu2019large}. Despite not storing any replay data, our method performs roughly in between BiC and LwF, though there is still a considerable (and expected) gap between our data-free approach and BiC. In summary, these results indicate that \textit{our method achieves State-of-the-Art performance for Data-Free Class-Incremental Learning}, and \textit{our method closes much of the performance gap between Data-Free Class-Incremental Learning and State-of-the-Art replay-based methods.}

\noindent
\textbf{Ablation Study - CIFAR-100 Benchmark}: We separate the components of our method to independently evaluate their effect on final performance, shown in Tables~\ref{tab:ablation-cifar100}. We first look at the effect of removing task balancing loss weighting. As previously reported~\citep{lee2019overcoming, Shin:2017,wu2019large}, this loss weighting has a significant effect on our performance. 
Next, we show that replacing the modified cross-entropy loss with standard cross-entropy loss cuts performance in half. Similarly, we show that ablating real and synthetic data distillation have the same effect. This indicates that 1) all three of these losses are crucial for our approach, and 2) we still establish SOTA performance despite removing any of these three losses. Finally, we see that removing the fine-tuning cross-entropy loss has the largest effect on performance. Conceptually, this makes sense because without this loss there is no way to distinguish new task classes from previous task classes.

\begin{table}[t]
\small
\caption{Results (\%) for class-incremental learning on five task ImageNet-50. A coreset of 2000 images is leveraged for replay-based methods, and thus \textit{these methods do not meet problem the DFCIL constraints}. Results are reported as a single run.}
\centering
\begin{tabular}{c | c | c }
    \hline
    Method & Replay Data & $A_{N}$ ($\uparrow$) \\
    \hline
    Upper Bound & None & $ 89.8 $  \\
    \hline
    LwF~\citep{li2016learning} & None & $ 19.4 $ \\
    Naive Rehearsal & Coreset & $ 78.9 $  \\
    LwF~\citep{li2016learning} & Coreset & $ \mathbf{ 84.8 } $ \\
    \hline 
    Ours & Synthetic & $ 71.5 $\\
    \hline
\end{tabular}
\label{tab:imagenet-50}
\vspace{-3mm}
\end{table}
\noindent
\textbf{Class-Incremental Learning with Replay Data - ImageNet Benchmark}: Finally, we use the ImageNet dataset~\citep{russakovsky2015imagenet} to demonstrate how our method performs on large scale 224x224x3 images. Following prior work~\citep{wu2019large}, we train with a 18-layer ResNet~\citep{he2016deep} for 100 epochs. The learning rate is set to 0.1 and is reduced by 10 after 30, 60, 80, and 90 epochs. We use a weight decay of 0.0001 and batch size of 128. We use the same class-shuffling seed as prior work~\citep{Rebuffi:2016,wu2019large} and report top-5 accuracy on ten tasks of 100 classes. We scale down to 20k coreset images used in the full ImageNet experiments to 2k, consistent with the relative number of classes. We also double the number of training steps used to train $\mathcal{F}$. Every other experiment detail is the same as the CIFAR-100 experiments. 

The results are given in Table~\ref{tab:imagenet-50}. Importantly, this experiment is significant because the number of parameters stored for replay (2000*224*224*3 = 3e8) \textit{far exceeds} the number of parameters \textit{temporarily} stored for synthesizing images (3.3e6). Despite requiring only 100 times fewer parameters to store, our method performs reasonably close to replay on this large-scale image experiment. We also far outperform LwF, which is the only DFCIL method to have been previously tried on large-scale ImageNet experiments. Additional experiments on the challenging Tiny-ImageNet dataset~\citep{le2015tiny}, which demonstrate the scalability of our method, are available in the Appendix (\ref{app:full_results}).
\section{Conclusions}

We show that existing class-incremental learning methods perform poorly when learning a new task with real training data and preserving past knowledge with synthetic distillation data. We then contribute a new method which achieves SOTA performance for data-free class-incremental learning, and is comparable to SOTA replay-based approaches. Our research vision is to \textit{eliminate the need for storing replay data} in class-incremental learning, enabling broad and practical applications of computer vision. An incremental learning solution which does not store data will provide immediate impact to computer vision applications such as reducing memory requirements for autonomous vehicles (which generate an inordinate amount of data), eliminating the need to transfer private medical data for medical imaging research collaboration (which is limited by strict legal protections), or removing the need to track private data for personal device user recommendation systems.
\section*{Acknowledgements}

This work is supported by Samsung Research America.

{\small
\bibliographystyle{ieee_fullname}
\bibliography{references}
}

\clearpage
\appendix
\section*{Appendix}
\setcounter{figure}{0}
\setcounter{table}{0}
\renewcommand{\thetable}{\Alph{table}}
\renewcommand{\thefigure}{\Alph{figure}}
\renewcommand\thesection{\Alph{section}}
\begin{figure*}[h]

    \centering
    \begin{subfigure}{0.48\textwidth}
        \centering
        
        \includegraphics[width = \textwidth]{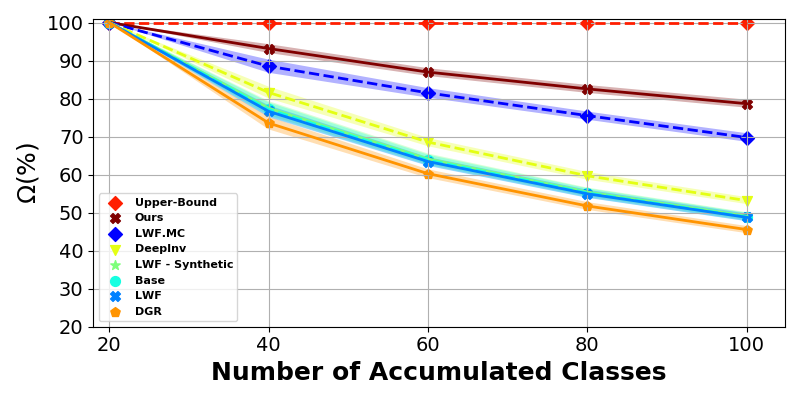}
        \caption{$\Omega$ curve for five task CIFAR-100 (without coreset)}
    \end{subfigure} \hfill
    \begin{subfigure}{0.48\textwidth}
        \centering
        
        \includegraphics[width = \textwidth]{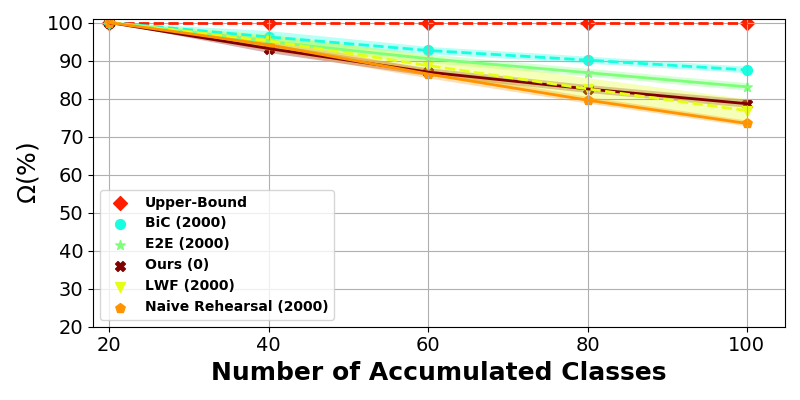}
        \caption{$\Omega$ curve for five task CIFAR-100 (with coreset)}
    \end{subfigure}
    
    \vspace{0.6cm}
    
    \begin{subfigure}{0.48\textwidth}
        \centering
        
        \includegraphics[width = \textwidth]{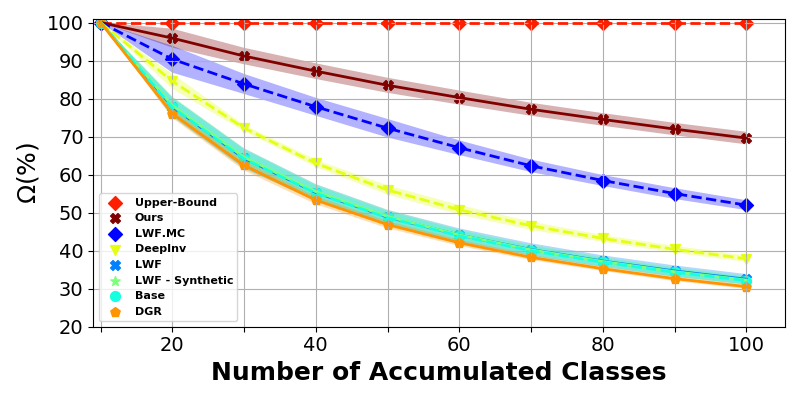}
        \caption{$\Omega$ curve for ten task CIFAR-100 (without coreset)}
    \end{subfigure} \hfill
    \begin{subfigure}{0.48\textwidth}
        \centering
        
        \includegraphics[width = \textwidth]{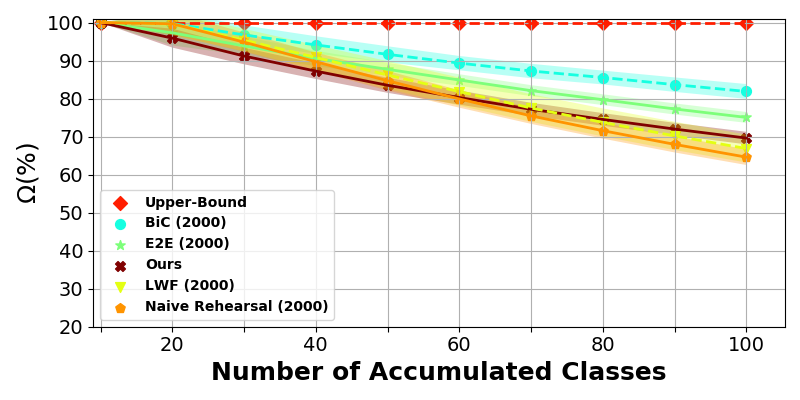}
        \caption{$\Omega$ curve for ten task CIFAR-100 (with coreset)}
    \end{subfigure}
    
    \vspace{0.6cm}
    
    \begin{subfigure}{0.48\textwidth}
        \centering
        
        \includegraphics[width = \textwidth]{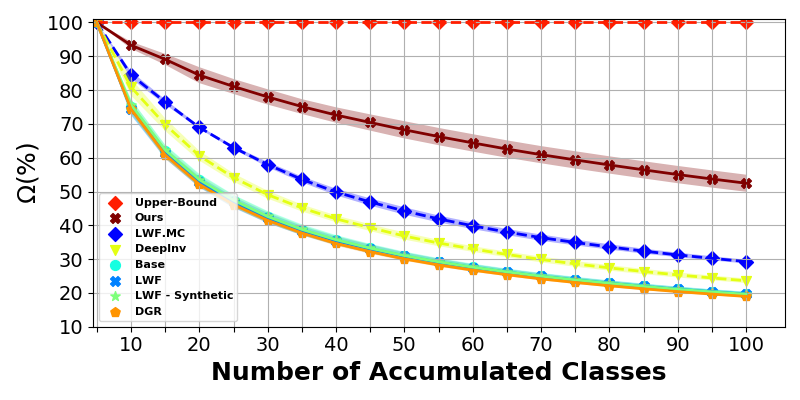}
        \caption{$\Omega$ curve for twenty task CIFAR-100 (without coreset)}
    \end{subfigure} \hfill
    \begin{subfigure}{0.48\textwidth}
        \centering
        
        \includegraphics[width = \textwidth]{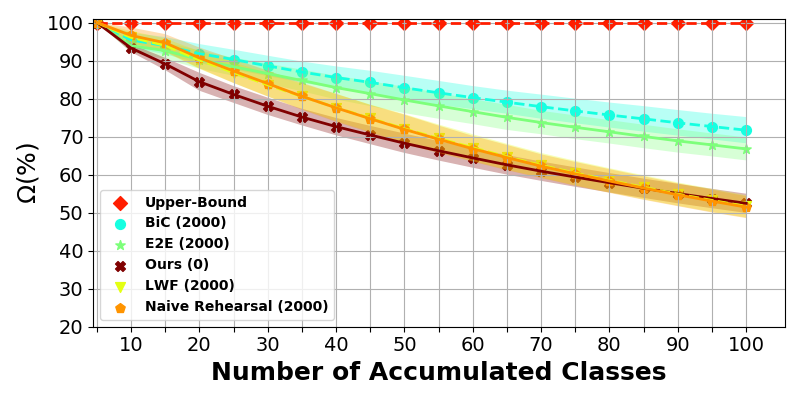}
        \caption{$\Omega$ curve for twenty task CIFAR-100 (with coreset)}
    \end{subfigure}
    
    \vspace{0.6cm}
    
    \begin{subfigure}{0.48\textwidth}
        \centering
        
        \includegraphics[width = \textwidth]{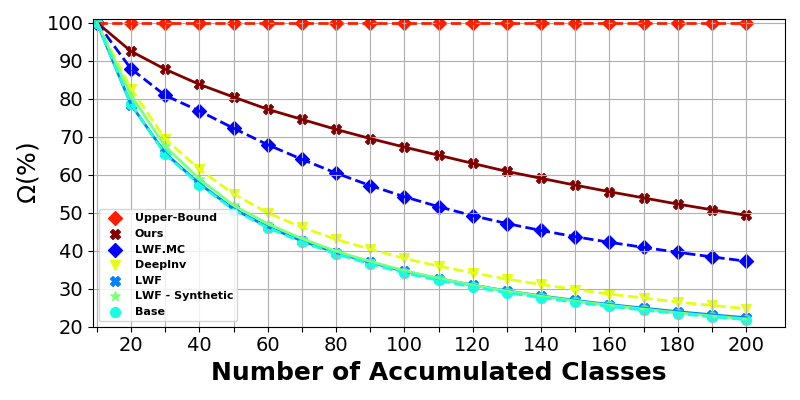}
        \caption{$\Omega$ curve for twenty task Tiny ImageNet (without coreset)}
    \end{subfigure} \hfill
    \begin{subfigure}{0.48\textwidth}
        \centering
        
        \includegraphics[width = \textwidth]{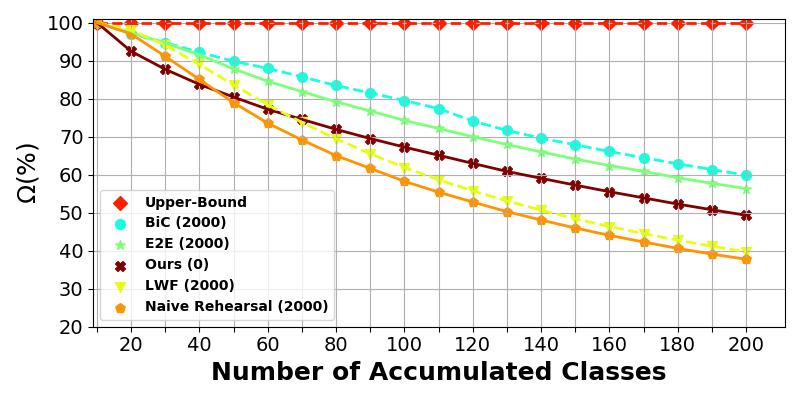}
        \caption{$\Omega$ curve for twenty task Tiny ImageNet (with coreset)}
    \end{subfigure}
    \vspace{0.5cm}
    \caption{$\Omega$ curves showing task number $t$ on the x-axis and $\Omega$ up to task $t$ on the y-axis.}
    \label{fig-app:omega}
\end{figure*}

\begin{table*}[t]
\small
\caption{Full Results (\%) for \textit{data-free} class-incremental learning on CIFAR-100 for various numbers of tasks (5, 10, 20). Results are reported as an average of 3 runs.}
\centering
\begin{tabular}{c | c | c c | c c | c c}
    \hline 
    \multicolumn{2}{c|}{Tasks} & \multicolumn{2}{c|}{5} & \multicolumn{2}{c|}{10} & \multicolumn{2}{c}{20} \\
    \hline
    Method & Replay Data & $A_{N}$ ($\uparrow$)  & $\Omega$ ($\uparrow$) & $A_{N}$ ($\uparrow$)  & $\Omega$ ($\uparrow$) & $A_{N}$ ($\uparrow$)  & $\Omega$ ($\uparrow$)\\
    \hline
    Upper Bound & None & $ 69.9 \pm 0.2 $ & $ 100.0 \pm 0.0 $ & $ 69.9 \pm 0.2 $ & $ 100.0 \pm 0.0 $ & $ 69.9 \pm 0.2 $ & $ 100.0 \pm 0.0 $ \\
    \hline
    Base & None & $ 16.4 \pm 0.4 $ & $ 48.9 \pm 1.1 $ & $ 8.8 \pm 0.1 $ & $ 32.1 \pm 1.1 $ & $ 4.4 \pm 0.3 $ & $ 19.7 \pm 0.7 $  \\ 
    LwF~\citep{li2016learning} & None & $ 17.0 \pm 0.1 $ & $ 49.5 \pm 0.1 $  & $ 9.2 \pm 0.0 $ & $ 33.3 \pm 0.9 $ & $ 4.7 \pm 0.1 $ & $ 20.1 \pm 0.3 $  \\ 
    LwF.MC~\citep{Rebuffi:2016} & None & $ 32.5 \pm 1.0 $ & $ 69.8 \pm 1.1 $  & $ 17.1 \pm 0.1 $ & $ 52.0 \pm 1.3 $ & $ 7.7 \pm 0.5 $ & $ 29.3 \pm 0.6 $  \\ 
    DGR~\citep{Shin:2017} & Generator & $ 14.4 \pm 0.4 $ & $ 45.5 \pm 0.9 $ & $ 8.1 \pm 0.1 $ & $ 30.5 \pm 0.6 $ & $ 4.1 \pm 0.3 $ & $ 19.0 \pm 0.3 $  \\
    LwF~\citep{li2016learning} & Synthetic & $ 16.7 \pm 0.1 $ & $ 49.8 \pm 0.1 $ & $ 8.9 \pm 0.0 $ & $ 32.3 \pm 0.0 $ & $ 4.7 \pm 0.0 $ & $ 19.7 \pm 0.0 $  \\
    DeepInversion~\citep{yin2020dreaming} & Synthetic & $ 18.8 \pm 0.3 $ & $ 53.2 \pm 0.9 $ & $ 10.9 \pm 0.6 $ & $ 37.9 \pm 0.8 $ & $ 5.7 \pm 0.3 $ & $ 23.6 \pm 0.7 $  \\ 
    \hline 
    Ours & Synthetic & $ \mathbf{43.9 \pm 0.9 } $ & $ \mathbf{ 78.6 \pm 1.1 } $ & $\mathbf{ 33.7 \pm 1.2 } $ & $ \mathbf{ 69.6 \pm 1.6 } $ & $ \mathbf{ 20.0 \pm 1.4 } $ & $ \mathbf{ 52.5 \pm 2.5 } $ \\
    \hline
\end{tabular}
\vspace{4mm}
\label{tab:results_cifar100_datafree_full}
\end{table*}
\begin{table*}[t]
\small
\caption{Results (\%) for class-incremental learning \textit{with replay data} on CIFAR-100 for various numbers of tasks (5, 10, 20). A coreset of 2000 images is leveraged for replay-based methods, and thus \textit{these methods do not meet problem the DFCIL constraints} (note we report for our method numbers \textit{without} any coreset). Results are reported as an average of 3 runs.}
\centering
\begin{tabular}{c | c | c c | c c | c c}
    \hline 
    \multicolumn{2}{c|}{Tasks} & \multicolumn{2}{c|}{5} & \multicolumn{2}{c|}{10} & \multicolumn{2}{c}{20} \\
    \hline
    Method & Replay Data & $A_{N}$ ($\uparrow$)  & $\Omega$ ($\uparrow$) & $A_{N}$ ($\uparrow$)  & $\Omega$ ($\uparrow$) & $A_{N}$ ($\uparrow$)  & $\Omega$ ($\uparrow$)\\
    \hline
    Upper Bound & None & $ 69.9 \pm 0.2 $ & $ 100.0 \pm 0.0 $ & $ 69.9 \pm 0.2 $ & $ 100.0 \pm 0.0 $ & $ 69.9 \pm 0.2 $ & $ 100.0 \pm 0.0 $ \\
    \hline
    Naive Rehearsal & Coreset & $ 34.0 \pm 0.2 $ & $ 73.4 \pm 0.8 $ & $ 24.0 \pm 1.0 $ & $ 64.6 \pm 2.1 $ & $ 14.9 \pm 0.7 $ & $ 51.4 \pm 2.9 $  \\ 
    LwF~\citep{li2016learning} & Coreset& $ 39.4 \pm 0.3 $ & $ 79.0 \pm 0.0  $ & $ 27.4 \pm 0.8 $ & $ 69.4 \pm 0.4 $ & $ 16.6 \pm 0.4 $ & $ 54.2 \pm 2.2 $  \\ 
    E2E~\citep{castro2018end} & Coreset & $ 47.4 \pm 0.8 $ & $ 83.1 \pm 1.0 $ & $ 38.4 \pm 1.3 $ & $ 75.0 \pm 1.4 $ & $ 32.7 \pm 1.9 $ & $ 66.8 \pm 3.0 $ \\
    BiC~\citep{wu2019large} & Coreset & $ \mathbf{53.7 \pm 0.4} $ & $ \mathbf{87.5 \pm 0.9} $ & $ \mathbf{45.9 \pm 1.8} $ & $ \mathbf{81.9 \pm 2.0} $ & $ \mathbf{37.5 \pm 3.2} $ & $ \mathbf{71.7 \pm 3.4} $  \\
    \hline 
    Ours & Synthetic & $ 43.9 \pm 0.9  $ & $ 78.6 \pm 1.1  $ & $ 33.7 \pm 1.2  $ & $  69.6 \pm 1.6  $ & $  20.0 \pm 1.4  $ & $ 52.5 \pm 2.5 $ \\
    \hline
\end{tabular}
\vspace{4mm}
\label{tab:results_cifar100_replay_full}
\end{table*}

\begin{table}[t]
\small
\caption{Results (\%) for \textit{data-free} class-incremental learning on Tiny ImageNet (20 tasks, 5 classes per task). Results are reported for a single run.}
\centering
\begin{tabular}{c | c | c c }
    \hline
    Method & Replay Data & $A_{N}$ ($\uparrow$)  & $\Omega$ ($\uparrow$) \\
    \hline
    Upper Bound & None & $ 55.5  $ & $ 100.0  $  \\ 
    \hline
    Base & None & $ 4.1  $ & $ 21.9  $  \\ 
    LwF~\citep{li2016learning} & None & $ 4.4  $ & $ 22.4  $  \\ 
    LwF.MC~\citep{Rebuffi:2016} & None & $ 8.8  $ & $ 37.2  $  \\ 
    LwF~\citep{li2016learning} & Synthetic & $ 4.0  $ & $ 22.0  $  \\ 
    DeepInversion~\citep{yin2020dreaming} & Synthetic & $ 5.1  $ & $ 24.8  $  \\ 
    \hline 
    Ours & Synthetic & $ 12.1  $ & $ 49.3  $  \\ 
    \hline
\end{tabular}
\vspace{4mm}
\label{tab:results_tinyimnet_datafree_full}
\end{table}
\begin{table}[t]
\small
\caption{Results (\%) for class-incremental learning \textit{with replay data} on Tiny ImageNet (20 tasks, 5 classes per task). A coreset of 2000 images is leveraged for replay-based methods, and thus \textit{these methods do not meet problem the DFCIL constraints} (note we report for our method numbers \textit{without} any coreset). Results are reported for a single run.}
\centering
\begin{tabular}{c | c | c c }
    \hline
    Method & Replay Data & $A_{N}$ ($\uparrow$)  & $\Omega$ ($\uparrow$) \\
    \hline
    Upper Bound & None & $ 55.5  $ & $ 100.0  $  \\ 
    \hline
    Naive Rehearsal & Coreset & $ 6.6  $ & $ 37.7  $  \\ 
    LwF~\citep{li2016learning} & Coreset & $ 6.9  $ & $ 39.7  $  \\ 
    E2E~\citep{castro2018end} & Coreset & $ 16.9  $ & $ 56.3  $  \\ 
    BiC~\citep{wu2019large} & Coreset & $ 17.4  $ & $ 59.8  $  \\ 
    \hline 
    Ours & Synthetic & $ 12.1  $ & $ 49.3  $  \\ 
    \hline
\end{tabular}
\vspace{4mm}
\label{tab:results_tinyimnet_replay_full}
\end{table}

\begin{table}[t]
\small
\caption{Results (\%) for class-incremental learning on five task ImageNet-50. A coreset of 2000 images is leveraged for replay-based methods, and thus \textit{these methods do not meet problem the DFCIL constraints}. Results are reported as a single run.}
\centering
\begin{tabular}{c | c | c }
    \hline
    Method & Replay Data & $A_{N}$ ($\uparrow$) \\
    \hline
    Upper Bound & None & $ 89.8 $  \\
    \hline
    LwF~\citep{li2016learning} & None & $ 19.4 $ \\
    LwF.MC~\citep{Rebuffi:2016} & None & $ 72.7 $ \\
    Naive Rehearsal & Coreset & $ 78.9 $  \\
    LwF~\citep{li2016learning} & Coreset & $ \mathbf{ 84.8 } $ \\
    \hline 
    Ours & Synthetic & $ 71.5 $\\
    \hline
\end{tabular}
\label{tab:imagenet-50_full}
\vspace{4mm}
\end{table}

\begin{table}[t]
\small
\caption{Range and chosen value of our hyperparameters, chosen with grid search}
\vspace{1mm}
\centering
\begin{tabular}{c|c|c}
    \hline
    Hyperparam. & Range & Value  \\
    \hline
    $\alpha_{con}$    & 1e-1, 1, 1e1 &  1 \\
    $\alpha_{div}$    & 1e-1, 1, 1e1 &  1 \\
    $\alpha_{stat}$   & 1, 1e1, 5e1, 1e2 &  5e1 \\
    $\alpha_{prior}$  & 1e-4, 1e-3, 1e-2, 1e-1, 1 & 1e-3 \\
    $\alpha_{temp}$   & 1, 1e1, 1e2, 1e3, 1e4 & 1e3 \\
    \hline
    $\lambda_{kd}$  & 1e-2, 1e-1, 1 & 1e-1 \\
    $\lambda_{ft}$  & 1e-2, 1e-1, 1 & 1e-1 \\
    \hline
\end{tabular}
\label{tab:param-search}
\vspace{12mm}
\end{table}

\begin{figure*}[t]
    \centering
    \begin{subfigure}[t]{0.45\textwidth}
        \centering
        \includegraphics[width=1\textwidth, trim = {0 0 0 0}, clip]{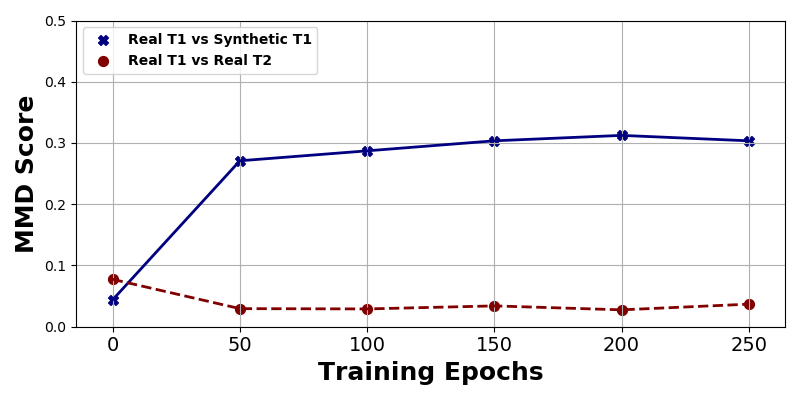}
        \caption{DeepInversion~\citep{yin2020dreaming}}
    \end{subfigure}
    \hspace{.5cm}
    \begin{subfigure}[t]{0.45\textwidth}
        \centering
        \includegraphics[width=1\textwidth, trim = {0 0 0 0}, clip]{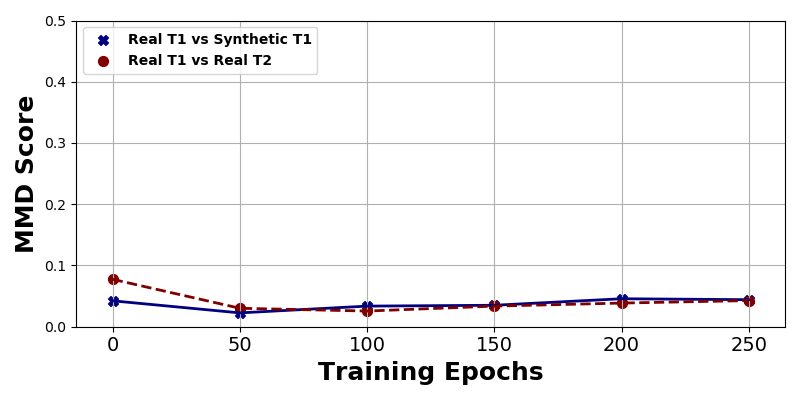}
        \caption{Our Method}
    \end{subfigure}
    \vspace{5mm}
    \caption{
    Maximum Mean Discrepancy (MMD) between feature embeddings of real task 1 data and synthetic task 1 data (blue), real task 2 data (red). Task 1 corresponds to ten classes of CIFAR-100 while task 2 corresponds to a different ten classes of CIFAR-100; the results are generated after training on task 2.
    }
    \label{fig:mmd}
\end{figure*}

\begin{figure*}[h!]
    \centering
    \begin{subfigure}[t]{0.35\textwidth}
        \centering
        \includegraphics[width=1\textwidth, trim = {0 0 0 0}, clip]{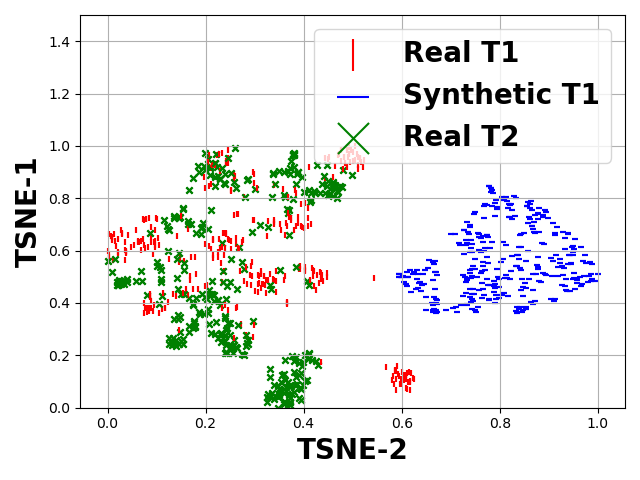}
         \caption{}
    \end{subfigure}
    \hspace{15mm}
    \begin{subfigure}[t]{0.35\textwidth}
        \centering
        \includegraphics[width=1\textwidth]{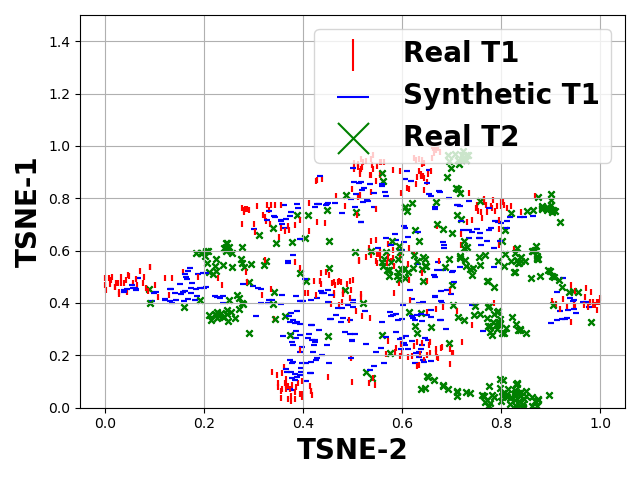}
        \caption{}
    \end{subfigure}
    \caption{t-SNE visualizations for (a) Figure 1.a (DeepInversion) and (b) Figure 1.c (Our Method) from the main text.}
    \label{fig-app:tsne}
\end{figure*}
\section{Additional Results}
\label{app:full_results}

In this section, we present additional experiments and results. For an alternative view of all results, we show $\Omega$ plotted by task in Figure~\ref{fig-app:omega}.

In Tables~\ref{tab:results_cifar100_datafree_full}/\ref{tab:results_cifar100_replay_full}, we expand our CIFAR-100 results with two additional methods: (1) LwF.MC~\citep{Rebuffi:2016}, a more powerful variant of LWF designed for class-incremental learning, and (2) End-to-End Incremental Learning~\citep{castro2018end} (E2E). In our implementation of E2E, we use the same data augmentations as our other experiments for a fair comparison. As previously published~\cite{wu2019large}, we see that E2E performs slightly worse than BiC and LwF.MC strongly outperforms LWF. Our approach consistently outperforms LwF.MC. 

We also report additional results on the Tiny-ImageNet dataset~\citep{le2015tiny} in Tables~\ref{tab:results_tinyimnet_datafree_full}/\ref{tab:results_tinyimnet_replay_full}, which contains 200 classes of 64x64 resolution images with 500 training images per class. We use the same experiment settings as CIFAR-100 with 10 classes per task and 20 tasks total. This is a highly challenging dataset with a low upper bound performance (drops from 69.9\% to 55.5\%), but we arrive at the same conclusions as we did for our CIFAR-100 experiments: our method outperforms all data-free class-incremental learning approaches, and performs slightly worse than state-of-the-art approaches which store 2000 images for replay. Importantly, the number of parameters stored for replay in these experiments (2000*64*64*3 = 2.5e7) \textit{far exceeds} the number of parameters \textit{temporarily} stored for synthesizing images (8.5e6). Note that this memory usage in our method can be completely removed at the cost of additional computation.  Despite requiring only 10 times fewer parameters to store (and not storing \textit{any} training data), our method performs reasonably close to state-of-the-art.

Finally, we expand the main paper results in Table~\ref{tab:imagenet-50_full} to include LWF.MC. Our method and LWF.MC perform similarly, indicating that more work is needed to scale our approach to large 224x224x3 images. This is not surprising because prior work~\citep{luo2020large} requires \textbf{1 generator per class} to scale data-free generative distillation up to ImageNet. We do not have the computational resources to perform this (e.g., full 1000 class ImageNet would require 1000 generators). Instead, our work demonstrates the need for generative data-free knowledge distillation to be \textit{efficiently} scaled up to the 224x224x3 images of ImageNet. We leave this to future work.  We kindly acknowledge that recent works which replay from a generator (close to our setting) also use small variants of ImageNet in their experiments~\citep{ayub2021eec,cong2020gan, ostapenko2019learning}.

\section{Additional Baseline Diagnosis with MMD}
\label{app:mmd}

In Section 5, we analyze \textit{representational distance} between embedded features with a metric that captures the distance between mean embedded images of two distribution samples. This metric is Mean Image Distance (MID) and is calculated with a reference sample of images $x_a$ and another sample of images $x_b$, where a \textit{high score} indicates \textit{dissimilar} features and a \textit{low score} indicates \textit{similar} features. In this section, we repeat the Section 5 experiments with the commonly used unbiased Maximum Mean Discrepancy (MMD)~\citep{gretton2012kernel}, which gives the distance between embeddings of two distributions in a reproducing kernel Hilbert space.

As done in Section 5, we start by training our model for the first two tasks in the ten-task CIFAR-100 benchmark. We calculate MMD between feature embeddings of real task 1 data and real task 2 data, and then we calculate MMD between feature embeddings of real task 1 data and synthetic task 1 data. The results are reported in Figure~\ref{fig:mmd}. For (a) DeepInversion, the MMD score between real task 1 data and synthetic task 1 data is significantly higher than the MMD score between real task 1 data and real task 2 data. As found in Section 5, this indicates that the embedding space prioritizes \textit{domain} over \textit{semantics}, which is detrimental because the classifier will learn the decision boundary between synthetic task 1 and real task 2, introducing great classification error with real task 1 images. For (b) our method, the MMD score between real task 1 data and synthetic task 1 data is much lower, indicating that our feature embedding prioritizes \textit{semantics} over \textit{domain}.

\section{Additional Experiment Details}
\label{app:exp_deets}

The majority of experiment details are listed in the main text (Section 7) and are dataset specific. Additionally: (i) we augment training data using standard augmentations such as random horizontal flips and crops, (ii) results were generated using a combination of Titan X and 2080 Ti GPUs, and (iii) synthesized images are sampled from $\mathcal{F}$ at each training step.

\section{Hyperparameter Sweeps}
\label{app:param_sweeps}

We tuned hyperparameters using a grid search. The hyperparameters were tuned using k-fold cross validation with three folds of the training data on only half of the tasks. We do not tune hyperparameters on the full task set because tuning hyperparameters with hold out data from all tasks may violate the principal of continual learning that states each task in visited only once~\citep{Ven:2019}. The results reported outside of this section are on testing splits (defined in the dataset).

\section{Discussion of Class Shuffling Seeds}
\label{app:seed}

Our results are slightly lower than reported in prior work~\citep{Rebuffi:2016,wu2019large} because we re-implemented each method in our benchmarking environment. A major difference between our implementation and these works is that, instead of using a fixed seed for a single class-order, we instead \textit{randomly shuffle} the class and task order for each experiment run. The class order has a significant effect on the end results, with our top performing class order resulting in performance similar to results reported in~\citep{wu2019large}. We argue that shuffling the class order gives a better representation of method performance while acknowledging both approaches (shuffling and not shuffling) have merit.

\section{t-SNE Visualization}
\label{app:tsne}

In Figure~\ref{fig-app:tsne}, we show real t-SNE visualizations which reasonably approximate Figure 1.a (DeepInversion) and Figure 1.c (Our Method) from the main text. Results are shown after training the second task in the ten-task CIFAR-100 benchmark. \textit{Importantly, the distilling $\theta_{1,1}$ model and the synthetic data are the same for both methods; only the loss functions are different}.

\begin{figure}[h!]
    \centering
    \includegraphics[width=0.4\textwidth, trim = {0 0 0 0}, clip]{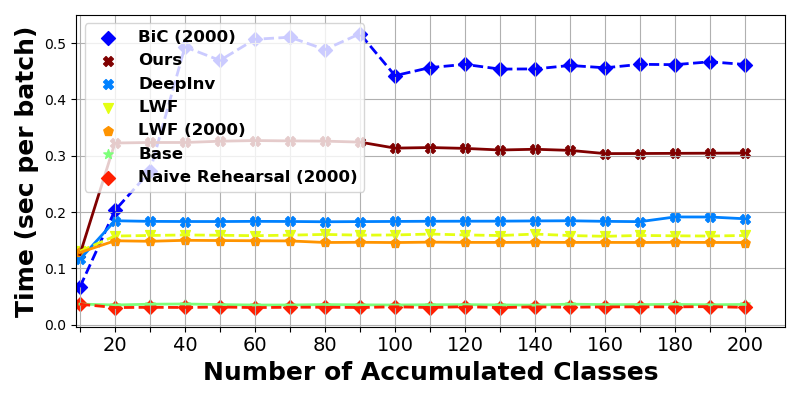}
    \vspace{2mm}
    \caption{Training time for the twenty task Tiny-ImageNet benchmark (Tables~\ref{tab:results_tinyimnet_datafree_full}/\ref{tab:results_tinyimnet_replay_full}).}
    \label{fig-app:time}
\end{figure}

\section{Training Time}
\label{app:time}

In Figure~\ref{fig-app:time}, we show the training time (seconds per training batch on a single Titain X Pascal GPU) for the twenty task Tiny-ImageNet benchmark (Tables~\ref{tab:results_tinyimnet_datafree_full}/\ref{tab:results_tinyimnet_replay_full}). Our method is faster than the SOTA replay-based method, BIC, yet slower than the other methods. All of these methods produce a model of the same architecture and therefore have the same inference time (except for BIC which has a \textit{very} small logit weighting operation).

\end{document}